\documentclass{article}

\usepackage{arxiv}

\usepackage[utf8]{inputenc} 
\usepackage[T1]{fontenc}    
\usepackage{hyperref}       
\usepackage{url}            
\usepackage{booktabs}       
\usepackage{amsfonts}       
\usepackage{nicefrac}       
\usepackage{microtype}      
\usepackage{amsthm}
\usepackage{amsmath}
\usepackage{cleveref}       
\usepackage{lipsum}         
\usepackage{graphicx}
\usepackage{natbib}
\usepackage{doi}
\usepackage{subfigure}
\usepackage{tabularray}

\theoremstyle{plain}
\newtheorem{theorem}{Theorem}[section]

\newtheorem{lemma}[theorem]{Lemma}

\theoremstyle{definition}
\newtheorem{definition}[theorem]{Definition}

\theoremstyle{remark}
\newtheorem{remark}[theorem]{Remark}

\title{Axiomatization of Gradient Smoothing in Neural Networks}

\newif\ifuniqueAffiliation
\uniqueAffiliationtrue

\ifuniqueAffiliation 
\author{ \href{https://orcid.org/0000-0003-3886-1300}{\includegraphics[scale=0.06]{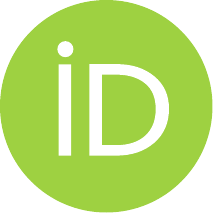}\hspace{1mm}Linjiang Zhou} \\
	School of Cyber Science and Engineering\\
	Wuhan University\\
	Hubei, China \\
	\texttt{linjiang@whu.edu.cn} \\
	\And
	\href{https://orcid.org/0000-0002-2044-0965}{\includegraphics[scale=0.06]{orcid.pdf}\hspace{1mm}Xiaochuan Shi} \\
	School of Cyber Science and Engineering\\
	Wuhan University\\
	Hubei, China \\
	\texttt{shixiaochuan@whu.edu.cn} \\
    \And
	\href{https://orcid.org/0000-0002-7443-6267}{\includegraphics[scale=0.06]{orcid.pdf}\hspace{1mm}Chao Ma} \\
	School of Cyber Science and Engineering\\
	Wuhan University\\
	Hubei, China \\
	\texttt{chaoma@whu.edu.cn} \\
    \And
	Zepeng Wang \\
	School of Cyber Science and Engineering\\
	Wuhan University\\
	Hubei, China \\
	\texttt{wangzepeng@whu.edu.cn} \\
}
\else
\usepackage{authblk}

\newbox{\orcid}\sbox{\orcid}{\includegraphics[scale=0.06]{orcid.pdf}} 
\author[1]{%
	\href{https://orcid.org/0000-0000-0000-0000}{\usebox{\orcid}\hspace{1mm}David S.~Hippocampus\thanks{\texttt{hippo@cs.cranberry-lemon.edu}}}%
}
\author[1,2]{%
	\href{https://orcid.org/0000-0000-0000-0000}{\usebox{\orcid}\hspace{1mm}Elias D.~Striatum\thanks{\texttt{stariate@ee.mount-sheikh.edu}}}%
}
\affil[1]{Department of Computer Science, Cranberry-Lemon University, Pittsburgh, PA 15213}
\affil[2]{Department of Electrical Engineering, Mount-Sheikh University, Santa Narimana, Levand}
\fi

\renewcommand{\shorttitle}{Axiomatization of Gradient Smoothing in Neural Networks}

\hypersetup{
pdftitle={Axiomatization of Gradient Smoothing in Neural Networks},
pdfsubject={XAI},
pdfauthor={Linjiang Zhou},
pdfkeywords={XAI},
}

\begin{document}
\maketitle

\begin{abstract}
	Gradients play a pivotal role in neural networks explanation. The inherent high dimensionality and structural complexity of neural networks result in the original gradients containing a significant amount of noise. While several approaches were proposed to reduce noise with smoothing, there is little discussion of the rationale behind smoothing gradients in neural networks. In this work, we proposed a gradient smooth theoretical framework for neural networks based on the function mollification and Monte Carlo integration. The framework intrinsically axiomatized gradient smoothing and reveals the rationale of existing methods. Furthermore, we provided an approach to design new smooth methods derived from the framework. By experimental measurement of several newly designed smooth methods, we demonstrated the research potential of our framework.
\end{abstract}

\keywords{Explainable Artificial Intelligence \and Neural Networks \and Gradient Smoothing}

\section{Introduction}\label{sec:intro}
Explanation for Artificial Intelligence (AI) is an inevitable part of AI applications with human interaction. For instance, explanation techniques are crucial in fields like medical image analysis \cite{alvarez2018towards}, financial data analysis \cite{zhou2023explainable}, and autonomous driving \cite{abid2022meaningfully}, where AI is applied to these data-sensitive and decision-sensitive fields. Additionally, given the prevalence of personal data protection laws in most countries and regions of the world, fully black-box AI models often face intense legal scrutiny \cite{doshi2017towards}.

After the development in recent years, some explanation methods try to explain the decisions of neural networks by the visualization of the decision basis and depiction of feature importance \cite{nauta2023anecdotal}. These local explanation methods aim to provide an explanation of neural networks on an individual sample. Moreover, the interpretation process or algorithm of these methods often utilizes the gradient of the neural network. For instance, Grad-CAM \cite{selvaraju2017grad}, Grad-CAM++ \cite{chattopadhay2018grad} and Score-CAM \cite{wang2020score} produces class activation map using gradients, and Integrate-Gradient \cite{sundararajan2017axiomatic} calculates integral gradients via sampling from gradients.

\begin{figure}
    \centering
    \subfigure[Input image]{
    \includegraphics[width=0.385\linewidth]{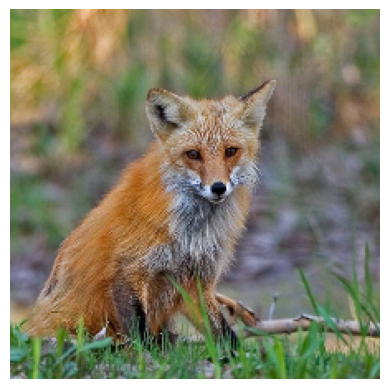}
    }
    \subfigure[The noisy original gradient]{
    \includegraphics[width=0.45\linewidth]{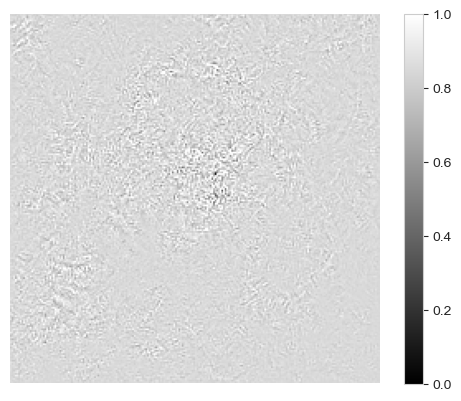}
    }
    \caption{An example of noisy original gradient in image classification model VGG16.}
    \label{fig:noisy}
\end{figure}

However, as noted in \cite{smilkov2017smoothgrad} and \cite{bykov2022noisegrad}, neural networks often have a significant amount of noise in their raw gradients due to their complex structure and high input dimensions. This noise might seriously impair interpretation. \cref{fig:noisy} shows an example of an original but noisy gradient. Therefore, \cite{smilkov2017smoothgrad} proposed a simple and effective method for smoothing the gradient called SmoothGrad. This method reduces noise by adding Gaussian noise to the samples multiple times and calculating the average of the gradients of these noise-added samples. \cite{bykov2022noisegrad} further expected to eliminate more noise by adding perturbations to the model parameters, called NoiseGrad.

As mentioned in \cite{ancona2017towards, nie2018theoretical, bykov2022noisegrad, nauta2023anecdotal}, the rationale by which the SmoothGrad and NoiseGrad methods reduce noise is currently unknown. Specifically, it is unclear how SmoothGrad effectively eliminates the majority of noise in the neural network gradient in a simple manner. That is, these explanation methods are inherently lacking in explainability.

So the relationship between the smoothed gradient and the raw gradient needs to be further studied. It is important to note that the principle behind the smoothed noise is unknown, therefore it cannot be guaranteed that the smoothed gradient is a fidelity representation of the original gradient. For instance, The research of \cite{adebayo2018sanity, kindermans2019reliability, yeh2019fidelity} has shown that many explanation methods do not accurately describe the real decisions of the model, but adopt some (in)fidelity approaches to improve the visualization performance.

Also, the rationale behind the phenomenon of gradient smoothing experiments also needs to be discussed. After adding Gaussian noise a sufficient number of times, it is unclear whether the smoothed gradient will converge in theory. The hyperparameter selection in gradient smoothing is based on empirical heuristics and lacks theoretical guidance.

Therefore, we aim to address these concerns by developing a comprehensive axiomatization framework of gradient smoothing, thus establishing a complete attribution chain of \textit{neural networks} - \textit{interpretable methods} - \textit{mathematical theorems}. By utilizing the function mollification theorem and Monte Carlo integration, we propose the Monte Carlo Gradient Mollification Formulation to present a theoretical explanation for gradient smoothing. The formulation uncovers the mathematical principles of gradient smoothing, which allows us to design a large variety of new gradient smoothing methods. Additionally, we present a straightforward and effective approach with mathematical principles and practical computational limitations. We utilized this approach to design five gradient smoothing kernel functions and examined the performance of smooth methods produced by these kernel functions on several metrics.

Our main contributions include:

\begin{itemize}
    \item We presented a comprehensive axiomatization framework of gradient smoothing, and revealed the basic mechanisms of the gradient smoothing methods.
    \item We proposed an approach for designing new gradient smoothing methods, and designed four new kernel functions for producing new gradient smoothing methods.
    \item The performance of these new methods has been experimentally explored using a range of evaluation metrics, demonstrating the research potential of this theoretical framework.
\end{itemize}

\section{Preliminary} \label{sec:pre}

In general, consider a neural network as a function $f(x;\theta): \mathbb{R}^D \to \mathbb{R}^C$, with the trainable parameters $\theta$. In the example of a classification function, the neural network will output a score for each class $c$, where $c\in \{0,1,\cdots,C\}$. When further considering only the output of the neural network in a single class $c$, the neural network can be simplified to a function $f^c(x;\theta): \mathbb{R}^D \to \mathbb{R}$, which maps the input $\mathbb{R}^D$ to the 1-dimension $\mathbb{R}$ space. For simplicity, we will use $f(x)$ to represent the neural network and only consider the output of the neural network on a single class.

The gradient of  $f(x)$ could be presented as,
\begin{equation}
    g(x)=\frac{\partial f(x)}{\partial x}
    \label{eq:1}
\end{equation}

The rationale for interpreting neural networks using original gradients, or using gradients multiplied by input values, is derived from Taylor's theorem. The first-order Taylor expansion on $f(x)$ is given by,
\begin{equation}
    f(x) = f(x_0)+ [\,g(x_0)]\,^\top(x-x_0)+R(x;x_0,\theta)
    \label{eq:2}
\end{equation}

If the residual term $R(x;x_0,\theta)$ is assumed to be a locally stable constant value, the gradient $g(x)$ could denote the contribution of each feature in $x$ to final outputs. Considering the local surroundings of a sample, neural networks tend not to present an ideal linearity but rather have a very rough and nonlinear decision boundary. [Need a Figure to explain] This complexity of neural networks and the input features usually leads to the assumption not being valid. Thus, this is a possible explanation for the large amount of noise in the original gradient. 

Recently, methods for reducing gradient noise can be roughly categorized into the following two groups:

\textbf{Adding noise to reduce noise.} SmoothGrad proposed by \cite{smilkov2017smoothgrad} introduces stochasticity to smooth the noisy gradients. SmoothGrad averages the gradients of random samples in the neighborhood of the input $x_0$. This could be described in,
\begin{equation}
\label{eq:3}
    sg(x)=\frac{1}{N}\sum_i^Ng( x+\varepsilon)
\end{equation}
where the sample times is $N$, and $\varepsilon$ is distributed as a Gaussian $\mathcal{N}(0,\sigma^2)$ with standard deviation $\sigma$. Similarly to SmoothGrad, NoiseGrad and FusionGrad presented in \cite{bykov2022noisegrad} additionally add perturbations to model parameters. The NoiseGrad could be defined as follows,
\begin{equation}
\label{eq:4}
    ng(x)=\frac{1}{M}\sum_i^Mg(x;\theta\cdot\eta)
\end{equation}
where the sample times is $M$, and $\eta$ follows distribution $\mathcal{N}(1,\sigma^2)$. And the FusionGrad is a mixup of NoiseGrad and SmoothGrad, which could be described in the same way,
\begin{equation}
\label{eq:5}
    fg(x)=\frac{1}{M}\frac{1}{N}\sum_j^M\sum_i^Ng(x+\varepsilon;\theta\cdot \eta)
\end{equation}
These simple methods are experimentally verified to be efficient and robust \cite{dombrowski2019explanations}.

\textbf{Improving backpropagation to reduce noise.} Deconvolution \cite{zeiler2014visualizing} and Guided Backpropagation \cite{springenberg2015striving} directly modifies the gradient computation algorithm of the ReLU function. Some other methods such as Feature Inversion \cite{du2018towards}, Layerwise Relevance Propagation \cite{bach2015pixel}, DeepLift \cite{shrikumar2017learning}, Occlusion \cite{ancona2017towards}, Deep Taylor \cite{montavon2017explaining}, etc. employ some additional features to approximate or improve the gradient for precise visualization. 

In the rest of this paper, we will focus on constructing a theoretical framework to explain the rationale for \textbf{adding noise to reduce noise}.

\section{Methodology}
The Monte Carlo gradient smoothing formulation will be introduced in this section. In order to facilitate the understanding of the derivation of the formula, we present it in a bottom-up manner. 
\subsection{Convolution and Dirac Function}\label{subsec:conv}

Convolution is an important functional operation. Referring to \cite{bak2010complex}, the convolution operation of function $f$ and $g$ could be defined as follows:

\begin{definition}
    \label{def:1}
    Let $f$ and $g$ be functions (generalized function). Symbol $*$ is defined as the convolution operator. The $f * g$ is defined by
    \begin{equation}
        (f*g)(x)=\int_{-\infty}^\infty f(t)g(x-t)\,dt
    \end{equation}
\end{definition}

Obviously, we can derive some useful lemmas:

\begin{lemma}
\label{lemma:1}
    If $f*g$ exists,
    \begin{equation}
        f*g=g*f
    \end{equation}
\end{lemma}

\begin{lemma}
\label{lemma:2}
    If $f*g$ exists and $f'$ and $g'$ exist,
    \begin{equation}
        (f*g)'=f'*g=f*g'
    \end{equation}
\end{lemma}
\begin{lemma}
\label{lemma:3}
 If $g$ is continuous, then $f*g$ is continuous.
\end{lemma}
The Dirac function is a generalized function and describes the limit of a set called the Dirac function sequence. In general, the 1-dimensional dirac function in $\mathbb{R}$ is defined as follows:
\begin{definition}
\label{def:2}
    The general definition of the Dirac function is,
    \begin{equation}
        \delta(x)=\begin{cases}
            +\infty & \text{if } x=0\\
            0 & \text{if } x\ne 0
        \end{cases}
    \end{equation}
    and it is constrained by,
    \begin{equation}
        \int_{-\infty}^\infty\delta(x)=1
    \end{equation}
\end{definition}

The Dirac function has some fundamental and important properties.
\begin{lemma}
\label{prop:1}
    If $f*\delta$ exists,
    \begin{equation}
        f*\delta = f
    \end{equation}
\end{lemma}

Using \cref{prop:1} and \cref{lemma:2}, we could easily get,
\begin{lemma}
\label{prop:2}
    If $f*\delta$ exists and $f'$ exist,
    \begin{equation}
        (f*\delta)' = f'*\delta = \delta * f' = f'
    \end{equation}
\end{lemma}
\cref{prop:1} and \cref{prop:2} also apply to function $f$ and Dirac functions $\delta$ in n-dimensions. All the proofs are in \cref{appendx:1}.

\subsection{Function Mollification}

Function mollification refers to the use of a mollifier to \textit{mollify} the target function so that the mollified function becomes \textit{continuous} or \textit{smooth}.
\begin{definition}\label{def:3}
    A function $\varphi(x): \mathbb{R}^n\to \mathbb{R}$ is a mollifier if it satisfies the following properties.
    \begin{enumerate}
        \item $\varphi(x)$ is of compact support,
        \item $\int_{\mathbb{R}^n} \varphi(x)\,dx=1$,
        \item $\lim_{\epsilon\to 0}\varphi_\epsilon(x):=\lim_{\epsilon\to 0}\epsilon^{-n}\varphi(\frac{x}{\epsilon})=\delta(x)$, where $\delta(x)$ is the Dirac function.
    \end{enumerate}
\end{definition}

The 3rd property in \cref{def:3} holds for almost all general functions consisting of primitive functions that satisfy the first two properties\footnote{The details could be found in \cite{Stein1971-qy} with Theorem 1.18.}. 

\begin{definition}\label{def:4}
    The convolution with a mollifier is called mollification,
    \begin{equation}
        f_\epsilon=f*\varphi_\epsilon
    \end{equation}
\end{definition}

By \cref{lemma:3}, if the mollifier $\varphi_\epsilon$ is continuous, $f_\epsilon$ is continuous. From \cref{lemma:3}, if $\varphi_\epsilon$ is continuous or locally continuous, $f_\varepsilon$ is also continuous or locally continuous. In other words, a mollifier $\varphi_\epsilon$ is a tool that smooths out the rough parts of a function $f$, similar to sandpaper with a granularity size of $1/\epsilon$. 

Noted that, in \cref{def:3}, $\varphi_\epsilon(x)$ defined a special function sequence, and the limit of the sequence is the Dirac function. Similarly, the definition of the Dirac sequence is presented as,
\begin{definition}
\label{def:5}
    if $\psi$ is absolute integrable function defined in $\mathbb{R}^n$ with $\int_{\mathbb{R}^n}\psi(x)\,dx=1$, the Dirac sequence is defined as,
    \begin{equation}
        \psi_\epsilon(x)=\epsilon^{-n}\psi(\frac{x}{\epsilon})
    \end{equation}
\end{definition}

In fact, mollification is a special case of convolution, and mollifier is a special case of Dirac sequences. By \cref{def:5}, $\varphi(x)$ defined a mollifier sequence $\{\varphi_\epsilon(x)\}$. So, $\varphi(x)$ is called kernel function because it determines the characteristic of the sequence $\{\varphi_\epsilon(x)\}$.

\subsection{Monte Carlo Gradient Mollification Formulation}

Given the noisy gradients and decision boundaries in most deep neural networks, we aim to smooth the neural networks with mollification. As introduced in \cref{sec:pre}, We simplify the neural networks into a function $f(x)$. With a proper mollifier $\varphi_\epsilon(x)$ given, we could use this mollifier to smooth the neural networks with mollification. Thus, by \cref{def:4}, we present the mollification of neural networks as,
\begin{equation}
    \label{eq:6}
    \begin{aligned}
        f_\epsilon(x) &= (f*\varphi_\epsilon)(x)\\
        &= \int_{\mathbb{R}^n}f(x-t)\varphi_\epsilon(t)\,dt\\
        &= \int_{\mathbb{R}^n}f(t)\varphi_\epsilon(x-t)\,dt
    \end{aligned}
\end{equation}
Accordingly, by \cref{lemma:2}, we could get the expression for the gradient of the smoothed neural network with mollification as,
\begin{equation}
    \label{eq:7}
    g_\epsilon(x)= (g*\varphi_\epsilon)(x) = \int_{\mathbb{R}^n}g(x-t)\varphi_\epsilon(t)\,dt
\end{equation}
Considering the computation complexity of the gradient, $f(t)$ and $g(t)$ are not easy to compute. So we chose to express $g_\epsilon(x)$ in the form of \cref{eq:7} instead of $\int_{\mathbb{R}^n}g(t)\varphi_\epsilon(x-t)\,dt$. And importantly, it is almost impossible to solve the analytic expression for $f_\epsilon$ in \cref{eq:6} and $g_\epsilon(x)$ in \cref{eq:7}. Therefore, $g_\epsilon(x)$ could only be solved numerically.

Note that $g_\epsilon(x)$ is essentially an integral and $x$ is a very high dimensional vector like $224\times224\times3$ in some image classification tasks. Therefore, Monte Carlo integration is almost the only efficient method for solving this problem. Thus, we give the Monte Carlo Gradient Mollification Formulation as follows,
\begin{theorem}
\label{thm:1}
    Given an instance $x_0$ and a $n$-dimension random variable $T$ obeying a distribution $P$ with probability density function $p(t)$, randomly sample $T$ for $N$ times to obtain a series of samples $t_i$. the Monte Carlo approximation of $g_\epsilon(x_0)$ is presented as,
    \begin{equation}
        \label{eq:8}
        mg_\epsilon(x_0)=\frac{1}{N}\sum_{i=1}^N\frac{g(x_0-t_i)\varphi_\epsilon(t_i)}{p(t_i)}
    \end{equation}
\end{theorem}

In \cref{thm:1}, $mg_\epsilon(x_0)$ is actually a statistic of the random variable $T$. We prove in the \cref{appendx:2} that this statistic is unbiased and consistent, which means that as $\lim_{N\to \infty}mg_\epsilon(x_0)=g_\epsilon(x_0)$.

\subsection{Axiomatization of Gradient Smoothing}\label{subsec:axio}

The existing gradient smoothing methods, SmooothGrad, NoiseGrad, and FusionGrad, are introduced in \cref{sec:pre}. Next, We aim to generalize these methods to \cref{thm:1}. That is, These methods are a special case of \cref{thm:1}, and they could all be derived from \cref{thm:1}.

\textbf{Derivation of SmoothGrad.}
In \cref{thm:1}, if $\varphi$ is the Probability Density Function (PDF) of a normal distribution, i.e. a Gaussian function, $\varphi$ satisfies all conditions in \cref{def:3}. So the Gaussian function is a reasonable mollifier. Next, set $\varphi_\epsilon=p$, that is, the PDF of $P$ is also $\varphi_\epsilon$. So SmoothGrad could be described as,
\begin{remark}
\label{remark:1}
    Given:
    \begin{equation}
    \label{eq:9}
        \begin{aligned}
            &\varphi_\epsilon(t) = \frac{1}{\sqrt{(2\pi)^n}\epsilon^{n}}\exp{-\frac{xx^\top}{2\epsilon^{2n}}}\\
            &p = \varphi_\epsilon
        \end{aligned}
    \end{equation}
    In \cref{thm:1}, $T\sim p(t)$, i.e. $T\sim \mathcal{N}(0,\epsilon^2)$,
    \begin{equation}
    \label{eq:10}
        mg_\epsilon(x_0)=\frac{1}{N}\sum_{i=1}^Ng(x_0-t_i).
    \end{equation}
    Thus, $ mg_\epsilon(x_0) = sg(x_0)$ in \cref{eq:3}. SmoothGrad $sg$ is a special case of $mg_\epsilon$ with condition \cref{eq:9}.
\end{remark}

\textbf{Derivation of NoiseGrad.}
With the same given condition of SmoothGrad, consider the parameter $\theta$ in neural networks as a smoothed target. 
\begin{remark}
\label{remark:2}
    Given:
    \begin{equation}
    \label{eq:11}
        \begin{aligned}
            &\varphi_\epsilon(t) = \frac{1}{\sqrt{(2\pi)^n}\epsilon^{n}\theta}\exp{-\frac{(x-\theta)(x-\theta)^\top}{2\epsilon^{2n}}}\\
            &p = \varphi_\epsilon
        \end{aligned}
    \end{equation}
    There we choose $T=\theta-\theta X, X\sim \mathcal{N}(0,\epsilon^2)$, because this makes $\theta-T=\theta X$ corresponding to $\theta\cdot \eta$ in \cref{eq:4}. So, in \cref{thm:1}, $T\sim p(t)$, i.e. $T\sim \mathcal{N}(\theta,\theta^2\epsilon^2)$,
    \begin{equation}
    \label{eq:12}
        mg_\epsilon(x_0)=\frac{1}{N}\sum_{i=1}^Ng(x_0;\theta+t_i).
    \end{equation}
    Thus, $ mg_\epsilon(x_0) = ng(x_0)$ in \cref{eq:3}. NoiseGrad $ng$ is a special case of $mg_\epsilon$ with condition \cref{eq:11}.
\end{remark}

\textbf{Derivation of FusionGrad.}
FusionGrad is a combination of SmoothGrad and NoiseGrad. It smooths the gradients with both input $x_0$ and parameters $\theta$. So it is easy to derive FusionGrad from \cref{thm:1}.
\begin{remark}
    \label{remark:3}
    Given:
    \begin{equation}
    \label{eq:13}
        \begin{aligned}
            &\varphi_\epsilon^{(x)}(t) = \frac{1}{\sqrt{(2\pi)^n}\epsilon^{n}\theta}\exp{-\frac{(x-\theta)(x-\theta)^\top}{2\epsilon^{2n}}}\\
            &p^{(x)} = \varphi_\epsilon^{(x)}\\
            &\varphi_\epsilon^{(\theta)}(t) = \frac{1}{\sqrt{(2\pi)^n}\epsilon^{n}\theta}\exp{-\frac{(x-\theta)(x-\theta)^\top}{2\epsilon^{2n}}}\\
            &p^{(\theta)} = \varphi_\epsilon^{(\theta)}
        \end{aligned}
    \end{equation}
    There mark $\varphi_\epsilon^{(x)}$ and $\varphi_\epsilon^{(\theta)}$ to distinguish the mollifiers for smoothing input $x_0$ and $\theta$. As described in \cref{remark:1} and \cref{remark:2}, we could get, in \cref{thm:1}, $T^{(x)}\sim p^{(x)}(t^{(x)})$ and $T^{(\theta)}\sim p^{(\theta)}(t^{(\theta)})$,
    \begin{equation}
    \label{eq:14}
        mg_\epsilon(x_0)=\frac{1}{N^{(\theta)}}\frac{1}{N^{(x)}}\sum_{i^{(\theta)}=1}^{N^{(\theta)}}\sum_{i^{(x)}=1}^{N^{(x)}}g(x_0+t^{(x)}_{i^{(x)}};\theta+t^{(\theta)}_{i^{(\theta)}}).
    \end{equation}
    Thus, $mg_\epsilon(x_0)=fg(x_0)$ in \cref{eq:5}. FusionGrad $fg$ is a special case of $mg_\epsilon$ with condition \cref{eq:13}.
\end{remark}

\section{Implications}

In this section, we will further discuss the implication of \cref{thm:1}. And since our study has uncovered a potentially infinite number of possible gradient smoothing methods, we expect to provide some useful guidelines for discovering potentially efficient gradient smoothing methods.

\subsection{Explanation for Gradient Smoothing}

The \cref{thm:1} reveals that the rational of \textbf{adding noise to reduce noise} is a Monte Carlo approximation of mollification for neural networks. 

\textit{1. As mentioned in \cref{lemma:3}, using continuous or locally continuous mollifer could make the smoothed function become continuous or locally continuous as well. }

\begin{figure}
    \centering
    \includegraphics[width=0.85\linewidth]{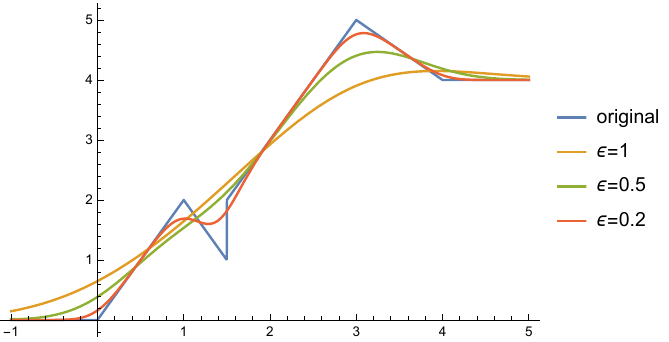}
    \caption{An example of mollification. A simple piecewise function $f(x):\mathbb{R}\to\mathbb{R}$ has a noncontinuous and noisy bound. And it was mollified by Gaussian kernel with different $\epsilon$. The details of the example can be found in \cref{appendx:3}}.
    \label{fig:toy}
\end{figure}

As shown in \cref{fig:toy}, the mollification could smooth out the rugged shape of the function, and the degree of smoothing is determined by $\epsilon$. The larger the value of $\epsilon$, the smoother the mollified function will be, and the smaller the value of $\epsilon$, the more the mollified function will retain its original undulations. As $\epsilon$ tends to 0, \cref{prop:2} shows the smoothed function will be the same as the original function. Thus, it answered the \textbf{RQ1} and so explained why gradient smoothing methods work.

\textit{2. The smoothed gradient converges as the number of samples increases, the limit of convergence is determined by the mollifier, and the speed of convergence is determined by the sampling distribution.}

In \cref{thm:1}, $mg_\epsilon$ will eventually converge to $g_\epsilon$ as the number of samples increases. The limit of convergence is only related to mollifier $\varphi_\epsilon(x)$. That is, after sampling enough times, the explanation performance of the smoothed gradient is determined by $\varphi_\epsilon(x)$. Thus, it answered the \textbf{RQ2} and means that we could get different smoothing methods by designing different $\varphi_\epsilon(x)$. And there are almost infinitely $\varphi_\epsilon(x)$ applicable in \cref{def:3}. Therefore, \cref{thm:1} reveals the existence of more possible research potential for gradient smoothing. 

\textit{3. The difference between the three methods, SmoothGrad, NoiseGrad, and FusionGrad, is that the input dimensions for smoothing are different, but they all employ a Gaussian kernel as the mollification kernel function.}

In detail, SmoothGrad smooths function $f(x;\theta)$ with $f(x;\theta)*\varphi_\epsilon$, NoiseGrad smooths function $f(\theta;x)$ with $f(\theta;x)*\varphi_\epsilon$, and FusionGrad smooths function $f(x,\theta)$. Moreover, FusionGrad smooths it in two times, that is $f(x,\theta) * \varphi_\epsilon^{(x)} * \varphi_\epsilon^{(\theta)}$. For simplicity, we will use SG, NG, and FG to denote these three smooth modes respectively.

\subsection{Approximation to Dirac Function}\label{subsec:app}
A large number of kernel functions satisfy \cref{def:3}, but not all of them are capable of efficiently smoothing neural networks. Therefore, we explore some mathematical and computational constraints and provide a simple approach to construct a suitable kernel function. We apply this approach to provide four kernel functions in addition to the Gaussian kernel function.

Based on some simple mathematical intuition \cite{febrianto2021mollified,timo2019mollify}, the kernel function generally selected needs to satisfy the following conditions,

\begin{equation}
\label{eq:24}
    \left\{\begin{array}{ll}
         \varphi(x) > 0 &, \text{for } \forall x \in \mathbb{R},\\
         \varphi(x)=\varphi(-x) &, \text{for } \forall x \in \mathbb{R}.
    \end{array}\right.
\end{equation}

To address the above limitations, we design a simple but effective method to quickly obtain the appropriate kernel function. Note that the indefinite integral of the Dirac function $\int_{-\infty}^x\delta(t)\,dt$ is a unit step function. That is, we can utilize an approximation of the Unit Step Function (USF) and then obtain an approximation of the Dirac function by solving for its derivatives. In the following, we construct five kernel functions including the Gaussian kernel using several common approximations of USF.

Gaussian kernel\footnote{Normal distribution could perform fast sampling with other methods. So there we do not present the inverse function $\Phi^{-1}(x)$.}:
\begin{equation}
\begin{array}{ll}
     &\Phi(x) = \frac{\text{Erf}(x/\sqrt{2})+1}{2}  \\
     & \varphi(x) = \frac{1}{\sqrt{2\pi}}e^{-\frac{x^2}{2}}
\end{array}
\end{equation}
Poisson kernel\footnote{This function is called Poisson kernel because it is one case of the Poisson formula for the solution of the Dirichlet problem in the upper half-plane}:
\begin{equation}
\begin{array}{lll}
     &\Phi(x) = \frac{\arctan\left(x\right)}{\pi }+\frac{1}{2}  \\
     &\Phi^{-1}(x)=\tan{(\frac{\pi}{2}(2x-1))}, 0<x<1 \\
     & \varphi(x) = \frac{1}{\pi  \left(x^2+1\right)}
\end{array}
\end{equation}
Hyperbolic kernel:
\begin{equation}
\begin{array}{lll}
     &\Phi(x) = \frac{\tanh (x)}{2}+\frac{1}{2}  \\
     &\Phi^{-1}(x) =  \text{arctanh}(2x-1), 0<x<1 \\
     & \varphi(x) = \frac{1}{2 \cosh ^2(x)}
\end{array}
\end{equation}
Sigmoid kernel:
\begin{equation}
\begin{array}{ll}
     &\Phi(x) = \frac{e^x}{e^x+1}  \\
     &\Phi^{-1}(x) = -\ln(\frac{1}{x}-1), 0<x<1\\
     & \varphi(x) = \frac{e^x}{\left(e^x+1\right)^2}
\end{array}
\end{equation}
Rectangular (Rect) kernel:
\begin{equation}
\begin{array}{ll}
     &\Phi(x) = \begin{cases}
         0&,x<-1\\
         \frac{1}{2}x+\frac{1}{2}&,-1\leq x \leq 1\\
         1&,x> 1
     \end{cases}  \\
     &\Phi^{-1}(x) = 2x-1, 0<x<1\\
     & \varphi(x) =\begin{cases}
         \frac{1}{2}&,|x|\leq1\\
         0&,\text{otherwise}
     \end{cases}
\end{array}
\end{equation}
\begin{figure}
    \centering
    \includegraphics[width=0.85\linewidth]{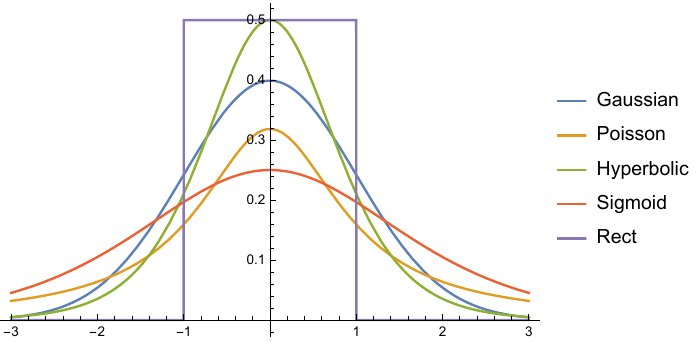}
    \caption{Visualization of Gaussian, Poisson, Hyperbolic, Sigmoid, Rect kernel with $\epsilon=1$}
    \label{fig:kernel}
\end{figure}

These kernel functions define the corresponding Dirac sequences. As shown in \cref{fig:kernel}, they have different shapes, which leads them to have different smoothing properties. In $n$-dimensions, assuming that the dimensions are independently and identically distributed, $\varphi(x) = \prod_i^n\varphi(x_i)$. And using variable substitution, it is also easy to obtain $\Phi_\epsilon^{-1}(x)=\Phi^{-1}(x)\times \epsilon$.

\subsection{Hyperparameter Selection}\label{subsec:hyper}
\begin{figure*}[tbh]
    \centering
    \includegraphics[width=0.75\textwidth]{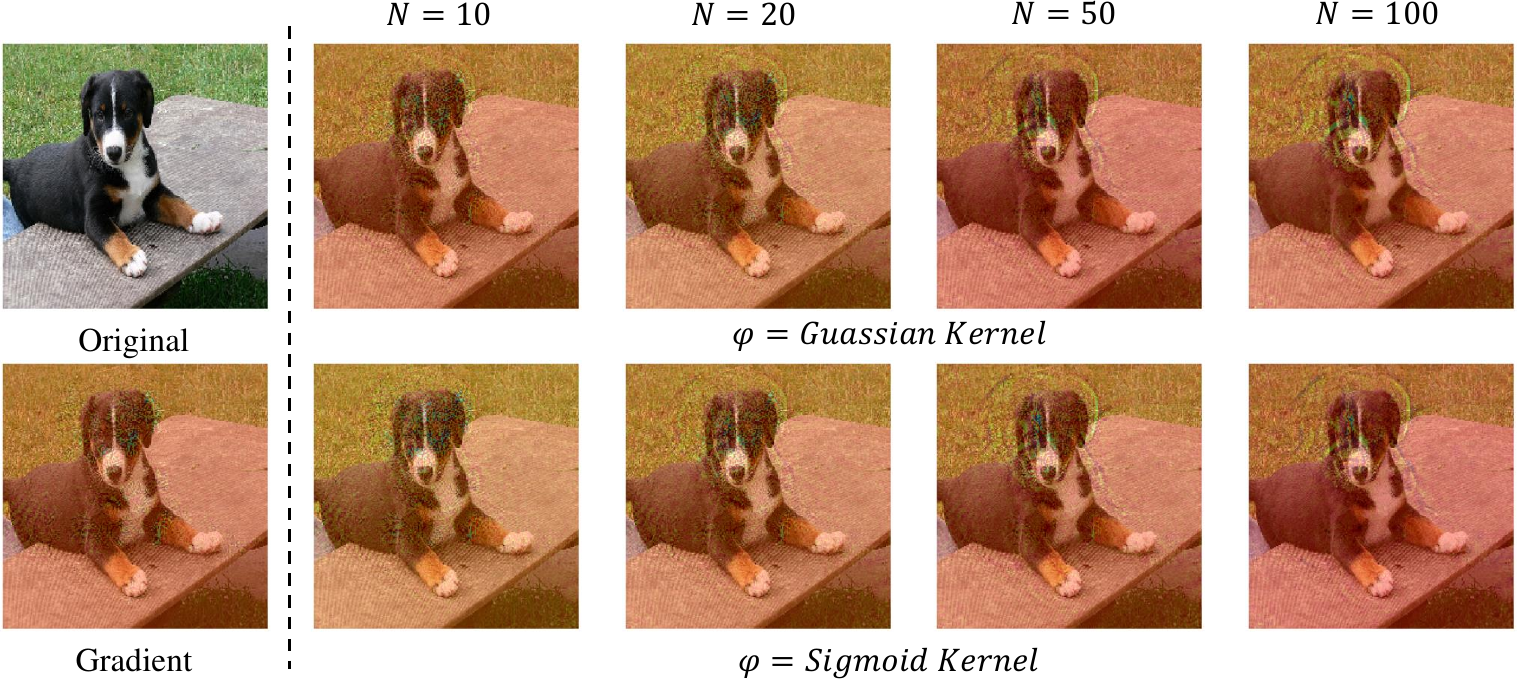}
    \caption{Visual explanation for random image labeled as \textit{dog} using SmoothGrad with Gaussian kernel and Sigmoid kernel. Visualization performance improves with varying numbers of samples, maintaining a convincing quality when reaching $N=50$.}
    \label{fig:dogn}
\end{figure*}

In our proposed smooth framework, there are two hyperparameters $N$ and $\epsilon$. From the perspective of Monte Carlo integration, the value of $N$ should be large enough for $mg$ to converge as much as possible. Combining the empirical values in \cite{smilkov2017smoothgrad, bykov2022noisegrad}, as shown in \cref{fig:dogn}, in practice SG has converged to a reasonable result at $N = 50$. Therefore, to balance performance and computation, we suggest that $N$ is taken to be around 50, and for FG then $N^{(\theta)}=M^{(x)}=50$.

With the constraints mentioned in \cref{subsec:app}, $\varphi_\epsilon$ is set the same as $p$. Therefore, $\varphi_\epsilon$ can be considered as a probability density function. This allows us to understand the selection of $\epsilon$ in a new sight, rather than on the basis of empirical values.
\begin{figure*}[tbh]
    \centering
    \includegraphics[width=0.75\textwidth]{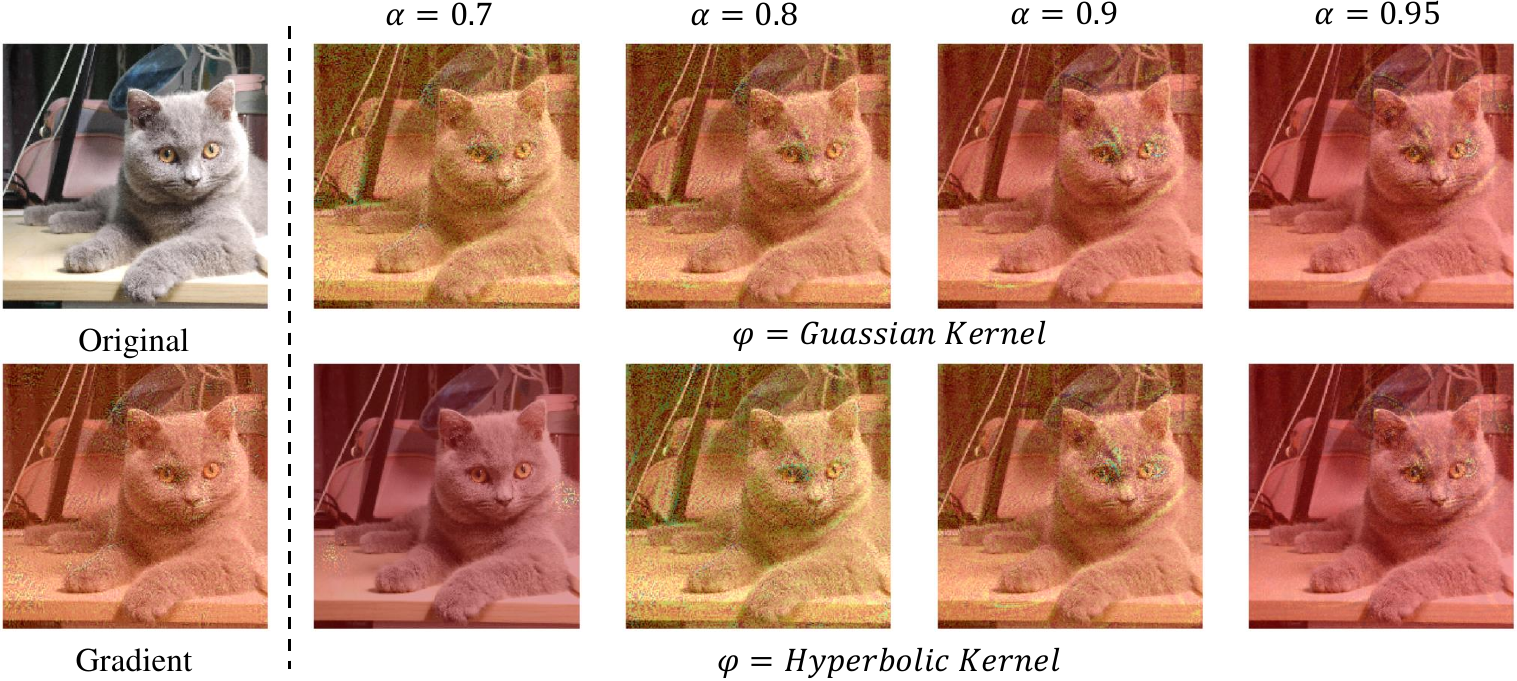}
    \caption{Visual explanation random image labeled as \textit{cat} using SmoothGrad with Gaussian kernel and Hyperbolic kernel. Visualization performance ranges with different $\alpha$, and when $\alpha=0.9$, the gradient map could present a decent performance, while when $\alpha=0.95$ it instead loses some details of the object edge.}
    \label{fig:cat_alpha}
\end{figure*}
Similar to confidence intervals and confidence levels, we expect that within a given interval $[-r,r]$, an appropriate $\epsilon$ will be selected such that the samples $t_i\sim \varphi_\epsilon(t)$ lies within that interval satisfies a given confidence level $\alpha$. Thus, $r$,$\epsilon$ and $\alpha$ satisfy the following equation,
\begin{equation}
    \label{eq:15}
    \frac{\alpha}{2}=\int_0^r\varphi_\epsilon(x)\,dx
\end{equation}

The value of $\epsilon$ could be calculated by substituting the given $r$ and $\alpha$ in \cref{eq:15}. The analytical solutions of $\epsilon$ for these five kernel functions are shown in \cref{appendx:4}. For SG, $r$ is suggested to be set to $r=max(x)-\bar{x}$, i.e. the maximum value of the dataset minus the mean value. For NG and FG, when adding noise to parameters of high layers in larger models, the slight disturbance could cause numerical calculation errors. Thus, referring to \cite{bykov2022noisegrad}, the $\sigma$ in \cref{eq:4} is set to $0.2$. Based on the $3\sigma$ rule of normal distribution, it is recommended to set $r=0.01$. So this makes it less likely to sample values that are outside the range of the dataset. According to statistical conventions, the value of $\alpha$ is recommended to be set to $0.9$ or $0.95$. In \cref{fig:cat_alpha}, we show the visualization of SG for different $\alpha$ values at $t=2$. Details of the hyperparameter settings and comparison with manual settings are shown in \cref{appendx:5}.

\begin{table*}
\centering
\caption{Performance comparison among 15 smoothing methods for three smoothing modes and five kernel function combinations. Original represents the original gradient. $\uparrow$ and $\downarrow$ represent whether these indicators are better higher or lower, respectively. The best-performing kernel function in smooth mode is marked in \textbf{bold}.}\label{table:1}
\resizebox{\linewidth}{!}{%
\begin{tblr}{
  cells = {c},
  cell{1}{1} = {r=2}{},
  cell{1}{2} = {c=3}{0.23\linewidth},
  cell{1}{5} = {c=3}{0.207\linewidth},
  cell{1}{8} = {c=3}{0.207\linewidth},
  cell{1}{11} = {c=3}{0.207\linewidth},
  cell{8}{2} = {c=3}{0.23\linewidth},
  cell{8}{5} = {c=3}{0.207\linewidth},
  cell{8}{8} = {c=3}{0.207\linewidth},
  cell{8}{11} = {c=3}{0.207\linewidth},
  vline{2,5,8,11} = {1,8}{},
  vline{2-13} = {2}{},
  vline{2-13} = {3-7}{},
  hline{1,3,8-9} = {-}{},
  hline{2} = {2-13}{},
}
Kernel & Consistency($\downarrow$) &  &  & Invariance($\uparrow$) &  &  & Localization($\uparrow$) &  &  & Sparseness($\uparrow$) &  & \\
 & SG & NG & FG & SG & NG & FG & SG & NG & FG & SG & NG & FG\\
Gaussian & 0.03234 & 0.02817 & 0.03070 & 0.5921 & 0.5031 & 0.5380 & \textbf{0.7040} & 0.6857 & \textbf{0.7085} & 0.5200 & \textbf{0.5659} & 0.3276\\
Poisson & \textbf{0.03031} & \textbf{0.02759} & \textbf{0.02881} & 0.5947 & \textbf{0.5056} & 0.2805 & 0.6747 & 0.6923 & 0.6325 & 0.5075 & 0.5324 & 0.2596\\
Hyperbolic & 0.03214 & 0.02821 & 0.03082 & 0.5966 & 0.5030 & 0.5386 & 0.6917 & 0.6867 & 0.7050 & 0.5195 & 0.5639 & 0.3268\\
Sigmoid & 0.03238 & 0.02826 & 0.03065 & 0.5921 & 0.5031 & 0.5382 & 0.6917 & 0.6940 & 0.6970 & \textbf{0.5202} & 0.5645 & 0.3243\\
Rect & 0.03211 & 0.02814 & 0.03069 & \textbf{0.5972} & 0.5034 & \textbf{0.5392} & 0.6930 & \textbf{0.6950} & 0.6965 & 0.5177 & 0.5624 & \textbf{0.3310}\\
Original & 0.02811 &  &  & 0.5016 &  &  & 0.6405 &  &  & 0.5587 &  & 
\end{tblr}
}
\end{table*}

\section{Experiment}
In this section, we test the performance of these five kernel functions in different explainability metrics.
\subsection{Experimental Settings}\label{subsec:setting}
\textbf{Metrics} The employment of solitary metrics to gauge gradient-based methods is a contentious issue \cite{nauta2023anecdotal}, and numerous current studies \cite{hedstrom2023quantus, yang2022psychological, liu2022rethinking, han2022explanation, jiang2021layercam} propose a plethora of evaluation metrics that even incorporate human evaluation. Nevertheless, we have chosen four properties to assess performance, as they align with the two fundamental objectives of model interpretation: reflecting model decisions \cite{adebayo2020debugging,sixt2020explanations,crabbe2020learning, fernando2019study, samek2016evaluating} and facilitating human understanding\cite{barkan2023visual,wu2022fam,ibrahim2022augmented, arras2022clevr,lu2021dance}.

\textit{Consistency} - Consistency, first introduced in \cite{adebayo2018sanity}, is primarily concerned with whether the XAI method is compatible with the model's learning capability. 

\textit{Invariance} - Invariance, proposed in \cite{kindermans2019reliability}, checks the XAI method to keep the output invariant in the case of constant data offsets in datasets with the same model architecture.

\textit{Localization} - Localization, employed in \cite{zhou2016learning, selvaraju2017grad, zhang2018top, wang2020score, barkan2023visual}, assesses the capability of the XAI method to accurately represent human-recognisable features in the input image.

\textit{Sparseness} - Sparseness, applied in \cite{chalasani2020concise}, measures whether the feature maps output by the XAI method are distinguishable and identifiable.

\textbf{Datasets and Models}
To apply these metrics for quantitative evaluation, referring to the set up in \cite{adebayo2018sanity} and \cite{kindermans2019reliability}, we chose MNIST \cite{LeCun2005TheMD} and CIFAR10 \cite{Krizhevsky2009LearningML} for experiments on Consistency and Invariance, and ILSVRC15 \cite{ILSVRC15} with target detection labels for experiments on Localization and Sparseness.

Correspondingly, we constructed MLP and CNN models for the Consistency and Invariance experiments. To measure the performance of the kernel function on models of different parameter sizes, we selected pre-trained VGG16 \cite{simonyan2014very}, MobileNet\cite{howard2017mobilenets}, and Inception\cite{szegedy2015going} for the Localization and Sparseness experiments.

\subsection{Results}

Based on the hyperparameter settings outlined in \cref{subsec:hyper}, we conducted experiments to test the performance of the five kernel functions. The aim of our study is not to identify a kernel function that completely outperforms existing methods but rather to explore the research potential of \cref{thm:1} and the possibility of designing new methods. This could help researchers to design better smoothing methods for different models and datasets based on \cref{thm:1}.

The experimental results are summarised in \cref{table:1}, showing the performance of 15 different smoothing methods for three smoothing modes and five combinations of kernel functions for different metrics. It could be observed that for Consistency and Invariance, the kernel functions Poisson and Rect perform better, while for Localization and Sparseness, the kernel functions Gaussian and Rect perform better. In general, we believe that Gaussian (which corresponds to existing methods like SmoothGrad), Poisson and Rect are convincing options for gradient smoothing. More qualitative and quantitative experimental results can be found in \cref{appendx:6}.

\section{Conclusion}
In this paper, we present a theoretical framework to explain the gradient smoothing approach. The existing gradient smoothing methods are axiomatized into a Monte Carlo smoothing formula, which reveals the mathematical nature of gradient smoothing and explains the rationale that enable these methods to remove gradient noise. Additionally, the theorem holds the potential for designing new smoothing methods. Therefore, we also suggest methods for designing smoothing approaches and propose several novel smoothing methods. Through experiments, we comprehensively measure the performance of these smoothing methods on different metrics and demonstrate the significant research potential of the theorem.

\textbf{Limitations and Future work}

\textit{Kernel function selection.} The selection of the kernel function should be related to the model parameters and dataset, and how to select the appropriate kernel function should be further explored.

\textit{Acceleration of Smooth Methods.} We only explored the selection of mollifer and ignored the sampling distribution. However, computing the gradient for NG and FG is time-consuming. Therefore, selecting a suitable sampling distribution to accelerate the calculation is a potential research topic.

\textit{Combination with other gradient-based methods.} SmoothGrad has been used to enhance explanation performance in Smooth Grad-CAM++\cite{omeiza2019smooth}, SS-CAM\cite{wang2020ss}, Smooth IG\cite{goh2021understanding}, etc. So the integration of this theoretical framework with other explanation approaches could be further explored.

\bibliographystyle{unsrtnat}
\bibliography{main}  

\begin{thebibliography}{57}
\providecommand{\natexlab}[1]{#1}
\providecommand{\url}[1]{\texttt{#1}}
\expandafter\ifx\csname urlstyle\endcsname\relax
  \providecommand{\doi}[1]{doi: #1}\else
  \providecommand{\doi}{doi: \begingroup \urlstyle{rm}\Url}\fi

\bibitem[Alvarez~Melis and Jaakkola(2018)]{alvarez2018towards}
David Alvarez~Melis and Tommi Jaakkola.
\newblock Towards robust interpretability with self-explaining neural networks.
\newblock \emph{Advances in neural information processing systems}, 31, 2018.

\bibitem[Zhou et~al.(2023)Zhou, Shi, Bao, Gao, and Ma]{zhou2023explainable}
Linjiang Zhou, Xiaochuan Shi, Yaxiong Bao, Lihua Gao, and Chao Ma.
\newblock Explainable artificial intelligence for digital finance and consumption upgrading.
\newblock \emph{Finance Research Letters}, 58:\penalty0 104489, 2023.

\bibitem[Abid et~al.(2022)Abid, Yuksekgonul, and Zou]{abid2022meaningfully}
Abubakar Abid, Mert Yuksekgonul, and James Zou.
\newblock Meaningfully debugging model mistakes using conceptual counterfactual explanations.
\newblock In \emph{International Conference on Machine Learning}, pages 66--88. PMLR, 2022.

\bibitem[Doshi-Velez and Kim(2017)]{doshi2017towards}
Finale Doshi-Velez and Been Kim.
\newblock Towards a rigorous science of interpretable machine learning.
\newblock \emph{arXiv preprint arXiv:1702.08608}, 2017.

\bibitem[Nauta et~al.(2023)Nauta, Trienes, Pathak, Nguyen, Peters, Schmitt, Schl{\"o}tterer, van Keulen, and Seifert]{nauta2023anecdotal}
Meike Nauta, Jan Trienes, Shreyasi Pathak, Elisa Nguyen, Michelle Peters, Yasmin Schmitt, J{\"o}rg Schl{\"o}tterer, Maurice van Keulen, and Christin Seifert.
\newblock From anecdotal evidence to quantitative evaluation methods: A systematic review on evaluating explainable ai.
\newblock \emph{ACM Computing Surveys}, 55\penalty0 (13s):\penalty0 1--42, 2023.

\bibitem[Selvaraju et~al.(2017)Selvaraju, Cogswell, Das, Vedantam, Parikh, and Batra]{selvaraju2017grad}
Ramprasaath~R Selvaraju, Michael Cogswell, Abhishek Das, Ramakrishna Vedantam, Devi Parikh, and Dhruv Batra.
\newblock Grad-cam: Visual explanations from deep networks via gradient-based localization.
\newblock In \emph{Proceedings of the IEEE international conference on computer vision}, pages 618--626, 2017.

\bibitem[Chattopadhay et~al.(2018)Chattopadhay, Sarkar, Howlader, and Balasubramanian]{chattopadhay2018grad}
Aditya Chattopadhay, Anirban Sarkar, Prantik Howlader, and Vineeth~N Balasubramanian.
\newblock Grad-cam++: Generalized gradient-based visual explanations for deep convolutional networks.
\newblock In \emph{2018 IEEE winter conference on applications of computer vision (WACV)}, pages 839--847. IEEE, 2018.

\bibitem[Wang et~al.(2020{\natexlab{a}})Wang, Wang, Du, Yang, Zhang, Ding, Mardziel, and Hu]{wang2020score}
Haofan Wang, Zifan Wang, Mengnan Du, Fan Yang, Zijian Zhang, Sirui Ding, Piotr Mardziel, and Xia Hu.
\newblock Score-cam: Score-weighted visual explanations for convolutional neural networks.
\newblock In \emph{Proceedings of the IEEE/CVF conference on computer vision and pattern recognition workshops}, pages 24--25, 2020{\natexlab{a}}.

\bibitem[Sundararajan et~al.(2017)Sundararajan, Taly, and Yan]{sundararajan2017axiomatic}
Mukund Sundararajan, Ankur Taly, and Qiqi Yan.
\newblock Axiomatic attribution for deep networks.
\newblock In \emph{International conference on machine learning}, pages 3319--3328. PMLR, 2017.

\bibitem[Smilkov et~al.(2017)Smilkov, Thorat, Kim, Vi{\'e}gas, and Wattenberg]{smilkov2017smoothgrad}
Daniel Smilkov, Nikhil Thorat, Been Kim, Fernanda Vi{\'e}gas, and Martin Wattenberg.
\newblock Smoothgrad: removing noise by adding noise.
\newblock \emph{arXiv preprint arXiv:1706.03825}, 2017.

\bibitem[Bykov et~al.(2022)Bykov, Hedstr{\"o}m, Nakajima, and H{\"o}hne]{bykov2022noisegrad}
Kirill Bykov, Anna Hedstr{\"o}m, Shinichi Nakajima, and Marina M-C H{\"o}hne.
\newblock Noisegrad—enhancing explanations by introducing stochasticity to model weights.
\newblock In \emph{Proceedings of the AAAI Conference on Artificial Intelligence}, volume~36, pages 6132--6140, 2022.

\bibitem[Ancona et~al.(2017)Ancona, Ceolini, {\"O}ztireli, and Gross]{ancona2017towards}
Marco Ancona, Enea Ceolini, Cengiz {\"O}ztireli, and Markus Gross.
\newblock Towards better understanding of gradient-based attribution methods for deep neural networks.
\newblock \emph{arXiv preprint arXiv:1711.06104}, 2017.

\bibitem[Nie et~al.(2018)Nie, Zhang, and Patel]{nie2018theoretical}
Weili Nie, Yang Zhang, and Ankit Patel.
\newblock A theoretical explanation for perplexing behaviors of backpropagation-based visualizations.
\newblock In \emph{International conference on machine learning}, pages 3809--3818. PMLR, 2018.

\bibitem[Adebayo et~al.(2018)Adebayo, Gilmer, Muelly, Goodfellow, Hardt, and Kim]{adebayo2018sanity}
Julius Adebayo, Justin Gilmer, Michael Muelly, Ian Goodfellow, Moritz Hardt, and Been Kim.
\newblock Sanity checks for saliency maps.
\newblock \emph{Advances in neural information processing systems}, 31, 2018.

\bibitem[Kindermans et~al.(2019)Kindermans, Hooker, Adebayo, Alber, Schütt, Dähne, Erhan, and Kim]{kindermans2019reliability}
Pieter-Jan Kindermans, Sara Hooker, Julius Adebayo, Maximilian Alber, Kristof~T Schütt, Sven Dähne, Dumitru Erhan, and Been Kim.
\newblock The (un) reliability of saliency methods.
\newblock \emph{Explainable AI: Interpreting, explaining and visualizing deep learning}, pages 267--280, 2019.

\bibitem[Yeh et~al.(2019)Yeh, Hsieh, Suggala, Inouye, and Ravikumar]{yeh2019fidelity}
Chih-Kuan Yeh, Cheng-Yu Hsieh, Arun Suggala, David~I Inouye, and Pradeep~K Ravikumar.
\newblock On the (in) fidelity and sensitivity of explanations.
\newblock \emph{Advances in Neural Information Processing Systems}, 32, 2019.

\bibitem[Dombrowski et~al.(2019)Dombrowski, Alber, Anders, Ackermann, M{\"u}ller, and Kessel]{dombrowski2019explanations}
Ann-Kathrin Dombrowski, Maximillian Alber, Christopher Anders, Marcel Ackermann, Klaus-Robert M{\"u}ller, and Pan Kessel.
\newblock Explanations can be manipulated and geometry is to blame.
\newblock \emph{Advances in neural information processing systems}, 32, 2019.

\bibitem[Zeiler and Fergus(2014)]{zeiler2014visualizing}
Matthew~D Zeiler and Rob Fergus.
\newblock Visualizing and understanding convolutional networks.
\newblock In \emph{Computer Vision--ECCV 2014: 13th European Conference, Zurich, Switzerland, September 6-12, 2014, Proceedings, Part I 13}, pages 818--833. Springer, 2014.

\bibitem[Springenberg et~al.(2015)Springenberg, Dosovitskiy, Brox, and Riedmiller]{springenberg2015striving}
J~Springenberg, Alexey Dosovitskiy, Thomas Brox, and M~Riedmiller.
\newblock Striving for simplicity: The all convolutional net.
\newblock In \emph{ICLR (workshop track)}, 2015.

\bibitem[Du et~al.(2018)Du, Liu, Song, and Hu]{du2018towards}
Mengnan Du, Ninghao Liu, Qingquan Song, and Xia Hu.
\newblock Towards explanation of dnn-based prediction with guided feature inversion.
\newblock In \emph{Proceedings of the 24th ACM SIGKDD International Conference on Knowledge Discovery \& Data Mining}, pages 1358--1367, 2018.

\bibitem[Bach et~al.(2015)Bach, Binder, Montavon, Klauschen, M{\"u}ller, and Samek]{bach2015pixel}
Sebastian Bach, Alexander Binder, Gr{\'e}goire Montavon, Frederick Klauschen, Klaus-Robert M{\"u}ller, and Wojciech Samek.
\newblock On pixel-wise explanations for non-linear classifier decisions by layer-wise relevance propagation.
\newblock \emph{PloS one}, 10\penalty0 (7):\penalty0 e0130140, 2015.

\bibitem[Shrikumar et~al.(2017)Shrikumar, Greenside, and Kundaje]{shrikumar2017learning}
Avanti Shrikumar, Peyton Greenside, and Anshul Kundaje.
\newblock Learning important features through propagating activation differences.
\newblock In \emph{International conference on machine learning}, pages 3145--3153. PMLR, 2017.

\bibitem[Montavon et~al.(2017)Montavon, Lapuschkin, Binder, Samek, and M{\"u}ller]{montavon2017explaining}
Gr{\'e}goire Montavon, Sebastian Lapuschkin, Alexander Binder, Wojciech Samek, and Klaus-Robert M{\"u}ller.
\newblock Explaining nonlinear classification decisions with deep taylor decomposition.
\newblock \emph{Pattern recognition}, 65:\penalty0 211--222, 2017.

\bibitem[Bak et~al.(2010)Bak, Newman, and Newman]{bak2010complex}
Joseph Bak, Donald~J Newman, and Donald~J Newman.
\newblock \emph{Complex analysis}, volume~8.
\newblock Springer, 2010.

\bibitem[Stein and Weiss(1971)]{Stein1971-qy}
Elias~M Stein and Guido Weiss.
\newblock \emph{Introduction to Fourier analysis on euclidean spaces ({PMS-32)}, volume 32}.
\newblock Princeton Mathematical Series. Princeton University Press, Princeton, NJ, November 1971.

\bibitem[Febrianto et~al.(2021)Febrianto, Ortiz, and Cirak]{febrianto2021mollified}
Eky Febrianto, Michael Ortiz, and Fehmi Cirak.
\newblock Mollified finite element approximants of arbitrary order and smoothness.
\newblock \emph{Computer Methods in Applied Mechanics and Engineering}, 373:\penalty0 113513, 2021.

\bibitem[Rohner(2019)]{timo2019mollify}
Timo Rohner.
\newblock Test functions, mollifiers and convolution.
\newblock \url{https://timorohner.com/files/mollifiers.pdf}, 2019.
\newblock Last accessed on 2024-01-27.

\bibitem[Hedstr{\"o}m et~al.(2023)Hedstr{\"o}m, Weber, Krakowczyk, Bareeva, Motzkus, Samek, Lapuschkin, and H{\"o}hne]{hedstrom2023quantus}
Anna Hedstr{\"o}m, Leander Weber, Daniel Krakowczyk, Dilyara Bareeva, Franz Motzkus, Wojciech Samek, Sebastian Lapuschkin, and Marina M-C H{\"o}hne.
\newblock Quantus: An explainable ai toolkit for responsible evaluation of neural network explanations and beyond.
\newblock \emph{Journal of Machine Learning Research}, 24\penalty0 (34):\penalty0 1--11, 2023.

\bibitem[Yang et~al.(2022)Yang, Folke, and Shafto]{yang2022psychological}
Scott Cheng-Hsin Yang, Nils Erik~Tomas Folke, and Patrick Shafto.
\newblock A psychological theory of explainability.
\newblock In \emph{International Conference on Machine Learning}, pages 25007--25021. PMLR, 2022.

\bibitem[Liu et~al.(2022)Liu, Li, Guo, Kong, Li, and Wang]{liu2022rethinking}
Yibing Liu, Haoliang Li, Yangyang Guo, Chenqi Kong, Jing Li, and Shiqi Wang.
\newblock Rethinking attention-model explainability through faithfulness violation test.
\newblock In \emph{International Conference on Machine Learning}, pages 13807--13824. PMLR, 2022.

\bibitem[Han et~al.(2022)Han, Srinivas, and Lakkaraju]{han2022explanation}
Tessa Han, Suraj Srinivas, and Himabindu Lakkaraju.
\newblock Which explanation should i choose? a function approximation perspective to characterizing post hoc explanations.
\newblock \emph{Advances in Neural Information Processing Systems}, 35:\penalty0 5256--5268, 2022.

\bibitem[Jiang et~al.(2021)Jiang, Zhang, Hou, Cheng, and Wei]{jiang2021layercam}
Peng-Tao Jiang, Chang-Bin Zhang, Qibin Hou, Ming-Ming Cheng, and Yunchao Wei.
\newblock Layercam: Exploring hierarchical class activation maps for localization.
\newblock \emph{IEEE Transactions on Image Processing}, 30:\penalty0 5875--5888, 2021.

\bibitem[Adebayo et~al.(2020)Adebayo, Muelly, Liccardi, and Kim]{adebayo2020debugging}
Julius Adebayo, Michael Muelly, Ilaria Liccardi, and Been Kim.
\newblock Debugging tests for model explanations.
\newblock \emph{Advances in Neural Information Processing Systems}, 33:\penalty0 700--712, 2020.

\bibitem[Sixt et~al.(2020)Sixt, Granz, and Landgraf]{sixt2020explanations}
Leon Sixt, Maximilian Granz, and Tim Landgraf.
\newblock When explanations lie: Why many modified bp attributions fail.
\newblock In \emph{International Conference on Machine Learning}, pages 9046--9057. PMLR, 2020.

\bibitem[Crabbe et~al.(2020)Crabbe, Zhang, Zame, and van~der Schaar]{crabbe2020learning}
Jonathan Crabbe, Yao Zhang, William Zame, and Mihaela van~der Schaar.
\newblock Learning outside the black-box: The pursuit of interpretable models.
\newblock \emph{Advances in neural information processing systems}, 33:\penalty0 17838--17849, 2020.

\bibitem[Fernando et~al.(2019)Fernando, Singh, and Anand]{fernando2019study}
Zeon~Trevor Fernando, Jaspreet Singh, and Avishek Anand.
\newblock A study on the interpretability of neural retrieval models using deepshap.
\newblock In \emph{Proceedings of the 42nd international ACM SIGIR conference on research and development in information retrieval}, pages 1005--1008, 2019.

\bibitem[Samek et~al.(2016)Samek, Binder, Montavon, Lapuschkin, and M{\"u}ller]{samek2016evaluating}
Wojciech Samek, Alexander Binder, Gr{\'e}goire Montavon, Sebastian Lapuschkin, and Klaus-Robert M{\"u}ller.
\newblock Evaluating the visualization of what a deep neural network has learned.
\newblock \emph{IEEE transactions on neural networks and learning systems}, 28\penalty0 (11):\penalty0 2660--2673, 2016.

\bibitem[Barkan et~al.(2023)Barkan, Asher, Eshel, Koenigstein, et~al.]{barkan2023visual}
Oren Barkan, Yuval Asher, Amit Eshel, Noam Koenigstein, et~al.
\newblock Visual explanations via iterated integrated attributions.
\newblock In \emph{Proceedings of the IEEE/CVF International Conference on Computer Vision}, pages 2073--2084, 2023.

\bibitem[Wu et~al.(2022)Wu, Chen, Che, and Pu]{wu2022fam}
Yuxi Wu, Changhuai Chen, Jun Che, and Shiliang Pu.
\newblock Fam: Visual explanations for the feature representations from deep convolutional networks.
\newblock In \emph{Proceedings of the IEEE/CVF Conference on Computer Vision and Pattern Recognition}, pages 10307--10316, 2022.

\bibitem[Ibrahim and Shafiq(2022)]{ibrahim2022augmented}
Rami Ibrahim and M~Omair Shafiq.
\newblock Augmented score-cam: High resolution visual interpretations for deep neural networks.
\newblock \emph{Knowledge-Based Systems}, 252:\penalty0 109287, 2022.

\bibitem[Arras et~al.(2022)Arras, Osman, and Samek]{arras2022clevr}
Leila Arras, Ahmed Osman, and Wojciech Samek.
\newblock Clevr-xai: A benchmark dataset for the ground truth evaluation of neural network explanations.
\newblock \emph{Information Fusion}, 81:\penalty0 14--40, 2022.

\bibitem[Lu et~al.(2021)Lu, Guo, Xing, and Noble]{lu2021dance}
Yang~Young Lu, Wenbo Guo, Xinyu Xing, and William~Stafford Noble.
\newblock Dance: Enhancing saliency maps using decoys.
\newblock In \emph{International Conference on Machine Learning}, pages 7124--7133. PMLR, 2021.

\bibitem[Zhou et~al.(2016)Zhou, Khosla, Lapedriza, Oliva, and Torralba]{zhou2016learning}
Bolei Zhou, Aditya Khosla, Agata Lapedriza, Aude Oliva, and Antonio Torralba.
\newblock Learning deep features for discriminative localization.
\newblock In \emph{Proceedings of the IEEE conference on computer vision and pattern recognition}, pages 2921--2929, 2016.

\bibitem[Zhang et~al.(2018)Zhang, Bargal, Lin, Brandt, Shen, and Sclaroff]{zhang2018top}
Jianming Zhang, Sarah~Adel Bargal, Zhe Lin, Jonathan Brandt, Xiaohui Shen, and Stan Sclaroff.
\newblock Top-down neural attention by excitation backprop.
\newblock \emph{International Journal of Computer Vision}, 126\penalty0 (10):\penalty0 1084--1102, 2018.

\bibitem[Chalasani et~al.(2020)Chalasani, Chen, Chowdhury, Wu, and Jha]{chalasani2020concise}
Prasad Chalasani, Jiefeng Chen, Amrita~Roy Chowdhury, Xi~Wu, and Somesh Jha.
\newblock Concise explanations of neural networks using adversarial training.
\newblock In \emph{International Conference on Machine Learning}, pages 1383--1391. PMLR, 2020.

\bibitem[LeCun and Cortes(2005)]{LeCun2005TheMD}
Yann LeCun and Corinna Cortes.
\newblock The mnist database of handwritten digits, 2005.
\newblock URL \url{https://api.semanticscholar.org/CorpusID:60282629}.
\newblock Last accessed on 2024-01-27.

\bibitem[Krizhevsky et~al.(2009)Krizhevsky, Hinton, et~al.]{Krizhevsky2009LearningML}
Alex Krizhevsky, Geoffrey Hinton, et~al.
\newblock Learning multiple layers of features from tiny images.
\newblock 2009.

\bibitem[Russakovsky et~al.(2015)Russakovsky, Deng, Su, Krause, Satheesh, Ma, Huang, Karpathy, Khosla, Bernstein, Berg, and Fei-Fei]{ILSVRC15}
Olga Russakovsky, Jia Deng, Hao Su, Jonathan Krause, Sanjeev Satheesh, Sean Ma, Zhiheng Huang, Andrej Karpathy, Aditya Khosla, Michael Bernstein, Alexander~C. Berg, and Li~Fei-Fei.
\newblock {ImageNet Large Scale Visual Recognition Challenge}.
\newblock \emph{International Journal of Computer Vision (IJCV)}, 115\penalty0 (3):\penalty0 211--252, 2015.
\newblock \doi{10.1007/s11263-015-0816-y}.

\bibitem[Simonyan and Zisserman(2014)]{simonyan2014very}
Karen Simonyan and Andrew Zisserman.
\newblock Very deep convolutional networks for large-scale image recognition.
\newblock \emph{arXiv preprint arXiv:1409.1556}, 2014.

\bibitem[Howard et~al.(2017)Howard, Zhu, Chen, Kalenichenko, Wang, Weyand, Andreetto, and Adam]{howard2017mobilenets}
Andrew~G Howard, Menglong Zhu, Bo~Chen, Dmitry Kalenichenko, Weijun Wang, Tobias Weyand, Marco Andreetto, and Hartwig Adam.
\newblock Mobilenets: Efficient convolutional neural networks for mobile vision applications.
\newblock \emph{arXiv preprint arXiv:1704.04861}, 2017.

\bibitem[Szegedy et~al.(2015)Szegedy, Liu, Jia, Sermanet, Reed, Anguelov, Erhan, Vanhoucke, and Rabinovich]{szegedy2015going}
Christian Szegedy, Wei Liu, Yangqing Jia, Pierre Sermanet, Scott Reed, Dragomir Anguelov, Dumitru Erhan, Vincent Vanhoucke, and Andrew Rabinovich.
\newblock Going deeper with convolutions.
\newblock In \emph{Proceedings of the IEEE conference on computer vision and pattern recognition}, pages 1--9, 2015.

\bibitem[Omeiza et~al.(2019)Omeiza, Speakman, Cintas, and Weldermariam]{omeiza2019smooth}
Daniel Omeiza, Skyler Speakman, Celia Cintas, and Komminist Weldermariam.
\newblock Smooth grad-cam++: An enhanced inference level visualization technique for deep convolutional neural network models.
\newblock \emph{arXiv preprint arXiv:1908.01224}, 2019.

\bibitem[Wang et~al.(2020{\natexlab{b}})Wang, Naidu, Michael, and Kundu]{wang2020ss}
Haofan Wang, Rakshit Naidu, Joy Michael, and Soumya~Snigdha Kundu.
\newblock Ss-cam: Smoothed score-cam for sharper visual feature localization.
\newblock \emph{arXiv preprint arXiv:2006.14255}, 2020{\natexlab{b}}.

\bibitem[Goh et~al.(2021)Goh, Lapuschkin, Weber, Samek, and Binder]{goh2021understanding}
Gary~SW Goh, Sebastian Lapuschkin, Leander Weber, Wojciech Samek, and Alexander Binder.
\newblock Understanding integrated gradients with smoothtaylor for deep neural network attribution.
\newblock In \emph{2020 25th International Conference on Pattern Recognition (ICPR)}, pages 4949--4956. IEEE, 2021.

\bibitem[Rudin(1987)]{walter1987real}
Walter Rudin.
\newblock \emph{Real and complex analysis, Third Edition}.
\newblock McGraw-Hill, 1987.

\bibitem[He et~al.(2016)He, Zhang, Ren, and Sun]{he2016deep}
Kaiming He, Xiangyu Zhang, Shaoqing Ren, and Jian Sun.
\newblock Deep residual learning for image recognition.
\newblock In \emph{Proceedings of the IEEE conference on computer vision and pattern recognition}, pages 770--778, 2016.

\bibitem[LeCun et~al.(1998)LeCun, Bottou, Bengio, and Haffner]{lecun1998gradient}
Yann LeCun, L{\'e}on Bottou, Yoshua Bengio, and Patrick Haffner.
\newblock Gradient-based learning applied to document recognition.
\newblock \emph{Proceedings of the IEEE}, 86\penalty0 (11):\penalty0 2278--2324, 1998.

\end{thebibliography}






\newpage
\appendix
\section{Proofs in \cref{subsec:conv}}\label{appendx:1}
The mathematical concepts mentioned in \cref{subsec:conv} are in the field of functional analysis \cite{walter1987real}. Thus for the convenience of readers to understand, we have provided all the proofs below.

Proof of \cref{lemma:1}.
\begin{proof}
    In \cref{def:1}, 
    \begin{equation}
        (f*g)(x)=\int_{-\infty}^\infty f(t)g(x-t)\,dt
    \end{equation}
    and using substitution, let $t = x - t'$. So we could get,
    \begin{equation}
        (f*g)(x)=\int_\infty^{-\infty} -f(x-t')g(t')dt'
    \end{equation}
    Next, exchange the upper and lower limits of integrals to obtain,
    \begin{equation}
        (f*g)(x)=\int_{-\infty}^\infty f(x-t')g(t')dt'
    \end{equation}
    In the form, we finally get $g*f$. Thus, \cref{lemma:1} is proved.
\end{proof}

Proof of \cref{lemma:2}.
\begin{proof}
    We compute the following
    \begin{equation}
        \begin{aligned}
            (f*g)'(x)&=\lim_{\Delta x\to 0}\frac{(f*g)(x+\Delta x) - (f*g)(x)}{\Delta x} \\
            &=\lim_{\Delta x\to 0}\frac{\int_{-\infty}^\infty f(y)g(x+\Delta x-y)\,dy - \int_{-\infty}^\infty f(y)g(x-y)\,dy}{\Delta x} \\
            &= \lim_{\Delta x\to 0} \int_{-\infty}^\infty f(y)\left(\frac{g(x+\Delta x-y)-g(x-y)}{\Delta x}\right) dy \\
            &= \int_{-\infty}^\infty f(y) \left(\lim_{\Delta x\to 0} \frac{g(x+\Delta x-y)-g(x-y)}{\Delta x} \right) dy \\
            &= \int_{-\infty}^\infty f(y)g'(x-y)dy \\
            &= (f*g')(x)
        \end{aligned}
    \end{equation}
    Next, using \cref{lemma:1}, we could get $(f*g)'=f*g'=f'*g$.
\end{proof}

Proof of \cref{lemma:3}
\begin{proof}
    If $g$ is continuous, and then for any convergent sequence $\{x_m\} \subset \mathbb{R}^n, x_m \to x \in \mathbb{R}^n $,
    \begin{equation}
        \lim_{m\to \infty} g(x_m)=g(x)
    \end{equation}
    Then, we could get
    \begin{equation}
    \begin{aligned}
        \lim_{m\to \infty}(f*g)(x_m)&=\lim_{m\to \infty}\int_{\mathbb{R}^n}f(y)g(x_m-y)\,dy\\
        &=\int_{\mathbb{R}^n}f(y)\lim_{m\to \infty}g(x_m-y)\,dy\\
        &=\int_{\mathbb{R}^n}f(y)g(x-y)\,dy\\
        &=(f*g)(x)
    \end{aligned}
    \end{equation}
    According to the definition of \textit{continuous}, $f*g$ is continuous.
\end{proof}
Proof of \cref{prop:1}.
\begin{proof}
    Using the \cref{def:2} and \cref{def:1} we could get
    \begin{equation}
        \begin{aligned}
            \int_{-\infty}^\infty \delta(y)f(x-y)\,dy &= \lim_{\epsilon\to 0}\int_{-\epsilon}^\epsilon \delta(y)f(x-y)\,dy \\
            &= f(x) \lim_{\epsilon\to 0}\int_{-\epsilon}^\epsilon\delta(y)dy \\
            &= f(x)
        \end{aligned}
    \end{equation}
    Thus, $\delta*f=f*\delta=f$
\end{proof}

Proof of \cref{prop:2}.
\begin{proof}
    \cref{prop:2} could be easily proved using \cref{lemma:2} and \cref{prop:1}.
\end{proof}

\section{Proofs of the Unbiasedness and Consistency }\label{appendx:2}
Proof of the unbiasedness of statistic $mg_\varepsilon(x_0)$. Unbiasedness means that the sample expectation of a statistic is the true value of the overall parameter.
\begin{proof}
According to the definition of mathematical expectations,
\begin{equation}
\begin{aligned}
    \mathbb{E}\left[mg_\varepsilon(x_0)\right]&=\mathbb{E}\left[\frac{1}{N}\sum_{i=1}^N\frac{g(x_0-t_i)\varphi_\epsilon(t_i)}{p(t_i)}\right]\\
    &=\frac{1}{N}\sum_{i=1}^N\mathbb{E}\left[\frac{g(x_0-t_i)\varphi_\epsilon(t_i)}{p(t_i)}\right]\\
    &=\frac{1}{N}\sum_{i=1}^N\int_{\mathbb{R}^n}\frac{g(x_0-t)\varphi_\epsilon(t)}{p(t)}p(t)\,dt\\
    &=\int_{\mathbb{R}^n}g(x_0-t)\varphi_\epsilon(t)\,dt\\
\end{aligned}
\end{equation}
By \cref{eq:7},
\begin{equation}
    \mathbb{E}\left[mg_\varepsilon(x_0)\right]=\int_{\mathbb{R}^n}g(x_0-t)\varphi_\epsilon(t)\,dt=g_\epsilon(x_0)
\end{equation}
\end{proof}
Proof of the consistency of statistic $mg_\varepsilon(x_0)$. Consistency means that as the number of samples increases, the estimates converge more and more to the true value. Specifically, the variance of the statistic converges to 0 as the sample size increases.
\begin{proof}
    According to the definition of variance,
    \begin{equation}
        \begin{aligned}
            \sigma^2\left[mg_\epsilon(x_0)\right]&=\sigma^2\left[\frac{1}{N}\sum_{i=1}^N\frac{g(x_0-t_i)\varphi_\epsilon(t_i)}{p(t_i)}\right]\\
            &=\frac{1}{N^2}\sum_{i=1}^N\sigma^2\left[\frac{g(x_0-t_i)\varphi_\epsilon(t_i)}{p(t_i)}\right]\\
        \end{aligned}
    \end{equation}
    Note that $G=\frac{g(x_0-T)\varphi_\epsilon(T)}{p(T)}$,
    \begin{equation}
            \sigma^2\left[mg_\epsilon(x_0)\right]=\frac{1}{N^2}\sum_{i=1}^N\sigma^2[G_i]=\frac{1}{N^2}(N\sigma^2[G])=\frac{\sigma^2[G]}{N}
    \end{equation}
    That is,
    \begin{equation}
        \sigma\left[mg_\epsilon(x_0)\right]=\frac{\sigma[G]}{\sqrt{N}}
    \end{equation}
    Clearly, the variance of statistic $mg_\epsilon(x_0)$ decreases as $N$ increases, i.e., $\lim_{N\to0}\sigma\left[mg_\epsilon(x_0)\right]=0$.
\end{proof}

\section{Details of the Mollification Example}
\label{appendx:3}
We construct a piecewise function $f(x): \mathbb{R} \to \mathbb{R}$ to simulate a rough and unsmooth function. The expression of $f(x)$ is,
\begin{equation}
    f(x)= \left\{\begin{array}{llllll}
         0\,&,\,x\leq0  \\
         2x\,&,\,0<x\leq1\\
         4-2x\,&,\,1\leq x<\frac{3}{2}\\
         -1+2x\,&,\,\frac{3}{2}\leq x<3\\
         8-x\,&,\,3\leq x<4\\
         4\,&,\,4\leq x
    \end{array}\right.
\end{equation}
The selected mollifier $\varphi_\epsilon(x)$ is,
\begin{equation}
    \varphi_\epsilon(x)=\frac{1}{\sqrt{2\pi}\epsilon}e^{-\frac{x^2}{2\epsilon^2}}
\end{equation}
The mollification $f*\varphi_\epsilon$ could be calculated as,
\begin{equation}
    \begin{aligned}f*\varphi_\epsilon(x)= &\frac{1}{2} (x-4) \text{erf}\left(\frac{x-4}{\sqrt{2} \epsilon }\right) -\frac{3}{2} (x-3) \text{erf}\left(\frac{x-3}{\sqrt{2} \epsilon }\right)+2 \text{erf}\left(\frac{x-1}{\sqrt{2} \epsilon }\right)\\
        &+x \left(-2 \text{erf}\left(\frac{x-1}{\sqrt{2} \epsilon }\right)+\text{erf}\left(\frac{x}{\sqrt{2} \epsilon }\right)+2 \text{erf}\left(\frac{2 x-3}{2 \sqrt{2} \epsilon }\right)\right)+\frac{5}{2} \text{erf}\left(\frac{3-2 x}{2 \sqrt{2} \epsilon }\right)\\
        &+\frac{\epsilon  \left(2 e^{-\frac{x^2}{2 \epsilon ^2}}+4 e^{-\frac{(3-2 x)^2}{8 \epsilon ^2}}+e^{-\frac{(x-4)^2}{2 \epsilon ^2}}-3 e^{-\frac{(x-3)^2}{2 \epsilon ^2}}-4 e^{-\frac{(x-1)^2}{2 \epsilon ^2}}\right)}{\sqrt{2 \pi }}+2 \sqrt{\frac{1}{\epsilon ^2}} \epsilon
    \end{aligned}
\end{equation}

\section{Details of Kernel Functions}
\label{appendx:4}
\begin{figure}[htp]
    \centering
    \includegraphics[width=0.6\textwidth]{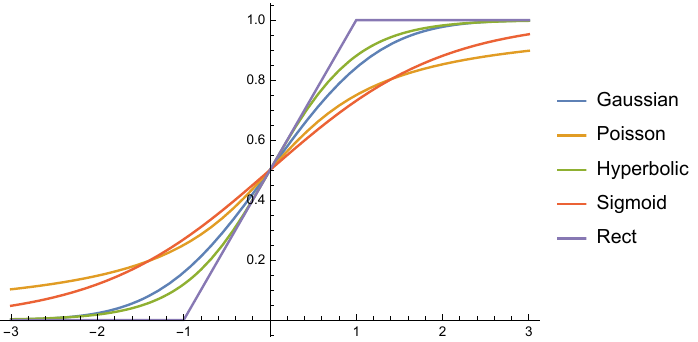}
    \caption{Visualization of the five USF approximation functions.}
    \label{fig:usf}
\end{figure}

The five kernel function is derived from USF approximation. \cref{fig:usf} shows images of these five USF approximation functions.

The USF approximation allows for quick computation of the analytical solution of $\epsilon$ in \cref{eq:15} when generating the kernel function. In contrast, the standard mollifier, as shown in \cref{eq:ap1}, usually adopted in the field of partial differential equations is difficult to get a simple format for Monte Carlo sampling. 

\begin{equation}
\label{eq:ap1}
    \varphi_\epsilon(x)= \begin{cases} \frac{1}{C}e^{(|x|^2-\epsilon)^{-1}} &,\text{if }|x|<\epsilon, \\ 
    0&,\text{if }|x|>\epsilon. \end{cases}
\end{equation}
The expressions for $\epsilon$ corresponding to these five kernel functions are given below,

Gaussian kernel:
\begin{equation}
    \epsilon = \frac{r}{\sqrt{2}\text{Erfinv}(\alpha)}
\end{equation}
There, $\text{Erfinv}$ is the inverse error function.

Poisson kernel:
\begin{equation}
    \epsilon = \frac{r}{\tan(\frac{\pi}{2} \alpha)}
\end{equation}

Hyperbolic kernel:
\begin{equation}
    \epsilon = \frac{r}{\arctan(\alpha)}
\end{equation}

Sigmoid kernel:
\begin{equation}
    \epsilon = \frac{r}{\ln(\frac{1+\alpha}{1-\alpha})}
\end{equation}

Rect kernel:
\begin{equation}
    \epsilon = \frac{2r}{\alpha}
\end{equation}

\section{Experiment for Hyperparameter Selection}
\label{appendx:5}

\begin{figure}[htp]
    \centering
    \includegraphics[width=0.6\textwidth]{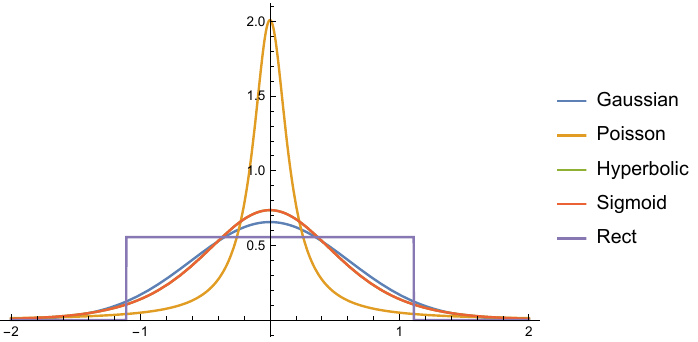}
    \caption{Visualization of five kernel functions with the same $\alpha$ and $r$ settings.}
    \label{fig:ap3}
\end{figure}

In \cref{subsec:hyper}, we provided an explanation for the hyperparameter selections in previous works. By \cref{eq:15}, a reasonable $\epsilon$ could be calculated with confidence level $\alpha$ and interval $[-r,r]$. \cref{fig:ap3} shows that in the same $\alpha$ and $r$ settings, the shape of these five kernel functions will tend to align, but still retain some differences, which causes the functions to have varying smoothing performance. In \cite{smilkov2017smoothgrad}, $\epsilon$ is recommended to set by,
\begin{equation}
    \epsilon = \beta \times (x_{min}-x_{max})
\end{equation}

Following the experimental setup in \cref{subsec:setting}, we tested the performance of the SG method using the Gaussian kernel function in different $\beta$. Our method is equivalent to $\beta$ value between 0.2 and 0.3. The experimental performance is shown in \cref{table:ap1}. The evaluation of Localization and Sparseness was only performed on VGG16. The performance difference among these hyperparameter settings is relatively slight. Our purpose is not to find an optimal hyperparameter but to show that our theoretical explanation of the settings of the hyperparameter $\epsilon$ in the gradient smoothing method is applicable.

\begin{table}[htp]
\centering
\caption{Performance comparison among different $\epsilon$ settings.}
\label{table:ap1}
\begin{tabular}{c|c|c|c|c}
\hline
$\epsilon$ value & Consistency($\downarrow$) & Invariance($\uparrow$) & Localization($\uparrow$)& Sparseness($\uparrow$) \\
\hline
$\beta=0.1$ & 0.03084 & 0.5217 & 0.699 & 0.5438 \\
$\beta=0.2$ & 0.03157 & 0.5584 & 0.694 & 0.5320 \\
$\beta=0.3$ & 0.03217 & 0.5922 & 0.698 & 0.5151 \\
ours($\beta=0.2\sim0.3$) & 0.03234 & 0.5922 & 0.699 & 0.5243 \\
\hline
original & 0.02811 & 0.5016 & 0.709 & 0.5393 \\
\hline
\end{tabular}
\end{table}

\section{More Experimental Details}
\label{appendx:6}

\subsection{Details of Metrics}
\textbf{Consistency}

In \cite{adebayo2018sanity}, the explanation methods are required to be consistent with the learning ability of the model. Similar to the statistical randomization test, the output of the explanation method should differentiate between models that are randomized and those that are trained normally. We refer to the work of \cite{adebayo2018sanity} and employ Spearman rank correlation to compute the difference between the outputs. Similarly, we use two regularisation methods to regularise the outputs, thus facilitating the computation of Spearman rank correlation. In \cref{table:1}, we show the average of the two methods.

\textbf{Invariance}

Proposed in \cite{kindermans2019reliability}, the explanation method was considered to maintain invariance to feature shifts that do not contribute. The authors applied a constant offset to the dataset to simulate invalid feature shifts. However, the authors only utilize visual analysis to determine whether the output of the explanation method remains invariant. Therefore, we further refer to the work of \cite{hedstrom2023quantus} and use rank correlation to quantify invariance.

\textbf{Localization}

Localization ability is a common evaluation metric for feature attribution-based explanation methods. Especially for a visual model-based explanation, this metric is optimized as almost the only goal by many explanation methods \cite{wu2022fam, jiang2021layercam, wang2020score}. However, it is important to note that many studies that measure localization ability based on object localization tasks tend to employ additional heuristics for generating bounding boxes for saliency maps. Therefore, based on the goals of our research, we do not select this approach, and instead, we employ the Point Game \cite{hedstrom2023quantus, zhang2018top} approach for evaluating localization ability. Specifically, the top-k (in our experiment, $k=5$) points with the maximal values on the output saliency map are selected, and we evaluate whether these points are in the bounding box of the ground truth or not. The accuracy in Point Game then quantitatively represents the localization ability.

\textbf{Sparseness}

In \cite{chalasani2020concise}, sparseness was introduced to evaluate the visualization capability of explanation methods. The idea of sparseness is that the explanation output should mainly focus on the important features, thus the saliency of the output should be as sparse as possible in order to make it easier for humans to visualize important features. Referring to \cite{chalasani2020concise}, we adopted the Gini index to quantify the sparsity of the output.

\subsection{Details of Experimental Environment}

VGG16, MobileNet, and Inception are constructed by pre-trained models released in Torchvision\footnote{https://pytorch.org/vision/stable/index.html}. The MLP architecture consists of two linear layers with 200 and 10 units respectively. The MLP was trained on MNIST by SGD optimizer with 20 epochs and the learning rate was set to 0.01. The CNN architecture contains a few layers as follows: [Conv(6), Maxpool(2), Conv(16), Maxpool(2), Linear(120), Linear(84), Linear(10)]. The CNN was trained on CIFAR10 by SGD optimizer with 20 epochs and the learning rate was set to 0.01. 

Due to the time-consuming computation of FG and NG (on average, it took 3-4 min to compute the gradient of one sample on our hardware), we randomly sampled 1,000 samples instead of all validation or test set samples for the comparison experiments. Our experiments were conducted on a machine with 4$\times$NVIDIA RTX 4090 and 2$\times$Intel Xeon Gold 6128. All experimental codes and saved model parameters with detailed results can be found in the Supplementary Material. 

\subsection{Additional Experimental Results}

For the comparison experiments on CNN, there are some differences in Consistency and Invariance metrics. Therefore, we present the detailed data separately in \cref{table:ap2}. It can be noticed that the values of Invariance metrics are significantly lower than the values in \cref{table:1}, and we speculate that this may be due to the low accuracy (0.5922 in our experiment) of the CNN model in the CIFAR10 dataset.

\begin{table}[htp]
\centering
\caption{Performance comparison among 15 smoothing methods of CNN in CIFAR10.}\label{table:ap2}
\begin{tblr}{
  cells = {c},
  cell{1}{1} = {r=2}{},
  cell{1}{2} = {c=3}{},
  cell{1}{5} = {c=3}{},
  cell{8}{2} = {c=3}{},
  cell{8}{5} = {c=3}{},
  vline{2,5} = {1,8}{},
  vline{2-7} = {2}{},
  vline{2-7} = {3-7}{},
  hline{1,3,8-9} = {-}{},
  hline{2} = {2-7}{},
}
Kernel & Consistency($\downarrow$) &  &  & Invariance($\uparrow$) &  & \\
 & SG & NG & FG & SG & NG & FG\\
Gaussian & 0.01721 & 0.01691 & 0.01574 & 0.02951 & 0.02720 & 0.02509\\
Poisson & 0.02042 & 0.01679 & 0.01838 & \textbf{0.07720} & 0.02077 & 0.02111\\
Hyperbolic & 0.01833 & 0.01722 & \textbf{0.01558} & 0.03096 & \textbf{0.02760} & \textbf{0.02604}\\
Sigmoid & 0.01745 & \textbf{0.01647} & 0.01743 & 0.02959 & 0.02745 & 0.02484\\
Rect & \textbf{0.01660} & 0.01677 & 0.01635 & 0.03022 & 0.02739 & 0.02600\\
Original & 0.01567 &  &  & 0.02622 &  & 
\end{tblr}
\end{table}

\begin{figure}[htp]
    \centering
    \includegraphics[width=0.9\textwidth]{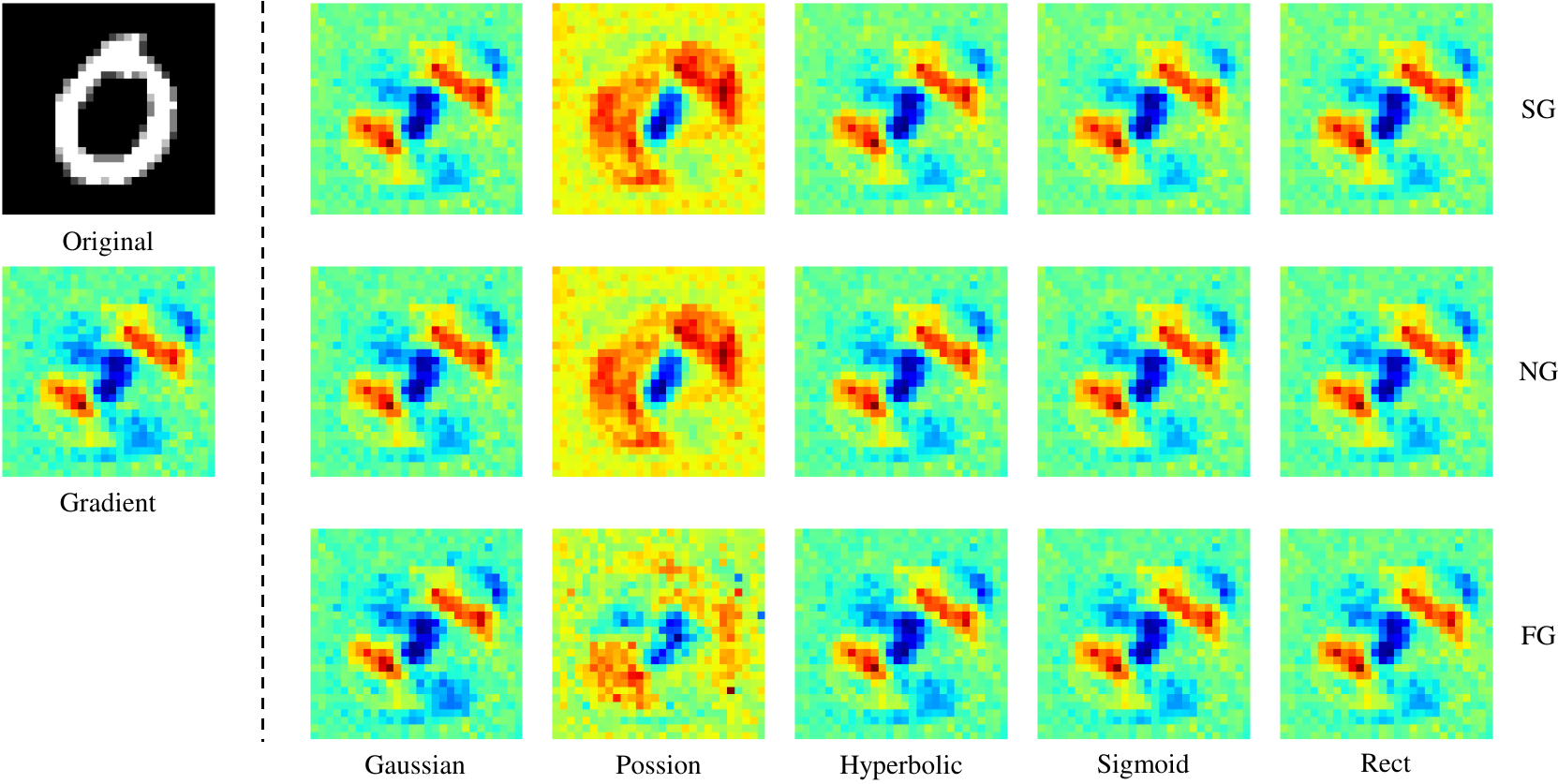}
    \caption{Visual explanation for a random image from MNIST labeled as \textit{0} with five kernel functions and three smoothing modes.}
    \label{fig:ap2}
\end{figure}

To visualize the difference in performance between these five kernel functions, we show an example of gradient smoothing for all five kernel functions on the MNIST dataset in \cref{fig:ap2}. It can be observed that the smoothing performance of the Poisson kernel function differs significantly from that of the other kernel functions. As shown in \cref{fig:ap3}, this difference is due to the distinct shape of the Poisson kernel function.

Localization is probably one of the most interesting metrics because it is directly related to the visualization of the gradient map. Therefore, we present the Localization metrics for the different gradient smoothing methods on the three models separately in Table \cref{table:ap3}. 

It is worth noting that the FG mode requires perturbing both model parameters and inputs. Therefore, in our experiments, we found that for models with larger parameters and deeper layers, the FG mode is extremely vulnerable to lead to numerical errors when perturbing the higher layers. Therefore, for models with fewer parameters such as VGG16 and MobileNet, the FG method was able to output valid gradient values, while for the Inception model, we dropped the anomalous data due to the presence of too many invalid gradients. Thus, in \cref{table:1} and \cref{table:ap3}, these outliers are excluded from our statistics.

In fact, in \cite{bykov2022noisegrad}, where FusionGrad is proposed, FusionGrad was only tested on ResNet18 \cite{he2016deep}, VGG, and LeNet \cite{lecun1998gradient}. The number of parameters in each of these models is relatively small. Therefore, we do not recommend using this method on models with a relatively large number of parameters.

\begin{table}[htp]
\centering
\caption{Localization performance comparison among 15 smoothing methods}
\label{table:ap3}
\begin{tblr}{
  cells = {c},
  cell{1}{1} = {r=2}{},
  cell{1}{2} = {c=3}{},
  cell{1}{5} = {c=3}{},
  cell{1}{8} = {c=3}{},
  cell{8}{2} = {c=3}{},
  cell{8}{5} = {c=3}{},
  cell{8}{8} = {c=3}{},
  vline{2,5,8} = {1,8}{},
  vline{2-10} = {2}{},
  vline{2-10} = {3-7}{},
  hline{1,3,8-9} = {-}{},
  hline{2} = {2-10}{},
}
Kernel & VGG16 &  &  & MobileNet   &  &  & Inception &  & \\
 & SG & NG & FG & SG & NG & FG & SG & NG & FG\\
Gaussian & 0.699 & 0.691 & 0.723 &\textbf{0.664} & 0.600 & \textbf{0.694} & 0.749 & 0.766 & -\\
Poisson & \textbf{0.712} & 0.693 & 0.707 & 0.552 & \textbf{0.624} & 0.558 & \textbf{0.760} & 0.760 & -\\
Hyperbolic & 0.701 & 0.695 & 0.722 & 0.636 & 0.612 & 0.688 & 0.738 & 0.753 & -\\
Sigmoid & 0.698 & \textbf{0.696} & \textbf{0.734} & 0.632 & 0.618 & 0.660 & 0.745 & 0.768 & -\\
Rect & 0.692 & 0.694 & 0.721 & 0.636 & 0.616 & 0.672 & 0.751 & \textbf{0.775} & -\\
Original & 0.709 &  &  & 0.572 &  &  & 0.761 &  & 
\end{tblr}

\end{table}
\end{document}